\documentclass{article}

\usepackage[numbers, square]{natbib} 
\usepackage{arxiv}

\usepackage{times}
\usepackage{url}
\usepackage{amsthm}
\usepackage{amsmath}
\usepackage{multirow}
\usepackage{array}
\usepackage{makecell}
\usepackage{txfonts}
\usepackage{lscape}
\usepackage{tabu}
\graphicspath{ {./figures/} }
\usepackage{graphicx}

\theoremstyle{definition}
\newtheorem{definition}{Definition}[section]

\def\realnumbers{\varmathbb{R}}
\def\complexnumbers{\varmathbb{C}}

\title{Evaluation of Complex-Valued Neural Networks on Real-Valued Classification Tasks}

\author{Nils M{\"o}nning \\
	Department of Computer Science\\
	University of York\\
	York, YO10 5GH, United Kingdom\\
	\texttt{nm819@york.ac.uk} \\
	\AND 
	Suresh Manandhar \\
	Department of Computer Science\\
	University of York\\
	York, YO10 5GH, United Kingdom\\
	\texttt{suresh.manandhar@york.ac.uk} \\
}

\begin{document}
\maketitle

\begin{abstract}
	Complex-valued neural networks are not a new concept, however, the use of real-valued models has often been favoured over complex-valued models due to difficulties in training and performance. When comparing real-valued versus complex-valued neural networks, existing literature often ignores the number of parameters, resulting in comparisons of neural networks with vastly different sizes. We find that when real and complex neural networks of similar capacity are compared, complex models perform equal to or slightly worse than real-valued models for a range of real-valued classification tasks. The use of complex numbers allows neural networks to handle noise on the complex plane. When classifying real-valued data with a complex-valued neural network, the imaginary parts of the weights follow their real parts. This behaviour is indicative for a task that does not require a complex-valued model. We further investigated this in a synthetic classification task. We can transfer many activation functions from the real to the complex domain using different strategies. The weight initialisation of complex neural networks, however, remains a significant problem.
\end{abstract}

\section{Introduction}
\label{intro}

In recent years complex-valued neural networks have been successfully applied to a variety of tasks, specifically in signal processing where the input data has a natural interpretation in the complex domain. Complex-valued neural networks are often compared to real-valued networks. We need to ensure that these architectures are comparable in their model size and capacity. This aspect of the comparison is rarely studied or only dealt with superficially. A metric for their capacity is the number of real-valued parameters. The introduction of complex numbers into a model increases the computational complexity and the number of real-valued parameters, but assumes a certain structure of weights and data input.

This paper explores the performance of complex-valued multi-layer perceptron (MLP) with varying depth and width. We consider the number of parameters and choice of activation function in benchmark classification tasks of real-valued data. We present a complex-valued multi-layer perceptron architecture and its training process. We consider various activation functions and the number of real-valued parameters in both the complex and real case. We propose two methods to construct comparable networks: 1) by setting a fixed number of real-valued neurons per layer and 2) by setting a fixed budget of real-valued parameters.
As benchmark task we choose MNIST digit classification  \cite{mnist:1998}, CIFAR-10 image classification \cite{cifar:2009}, CIFAR-100 image classification \cite{cifar:2009}, Reuters newswire topic classification (\textit{Reuters-21578, Distribution 1.0}). We use classification of synthetic data for further investigation.

\section{Related Literature}
\label{literature}
Complex-valued neural networks were first formally described by \citet{clarke:1990}. Several authors have since proposed complex versions of the backpropagation algorithm based on gradient descent \cite{benvenuto:1992, georgiou:1992, nitta:1993}. Inspired by work  on multi-valued threshold logic \cite{aizenberg:1977} from the 1970s, a multi-valued neuron and neural network was defined by Aizenberg et al. \cite{aizenberg:1992, aizenberg:2007} who also extends this idea to quaternions. In the 2000s, complex neural networks were successfully applied to a variety of tasks \cite{park:2002, goh:2006, ozbay:2008, hirose:2002}. These tasks mainly involved the processing and analysis of complex-valued data or data with an intuitive mapping to complex numbers. Particularly, images and signals in their wave form or Fourier transformation were used as input data to complex-valued neural networks \cite{hirose:2009}.

Another natural application of complex numbers are convolutions \cite{bruna:2015} which are used in image and signal processing. While real convolutions are widely used in deep learning for image processing, it is possible to replace them with complex convolutions \cite{trabelsi:2017, guberman:2016, popa:2017, haensch:2010}.

The properties of complex numbers and matrices can be used to define constraints on deep learning models. Introduced by \citet{arjovsky:2015}, and further developed by \citet{wisdom:2016}, complex-valued recurrent networks, that constrain their weights to be unitary matrices, reduce the impact of vanishing or exploding gradients. 

More recently, complex-valued neural networks have been used to learn filters as embeddings of images and audio signals \cite{trouillon:2016, sarroff:2015, drude:2016}. In addition, tensor factorisation has been applied to complex embeddings to predict the edges between entities of knowledge bases \cite{trouillon:2017}.

Despite their successes, complex neural networks have been less popular than their real-valued counter-parts. Potentially, because the training process and architecture design are less intuitive, which stems from stricter requirements for the differentiability of activation functions in the complex plane \cite{zimmermann:2011, hirose:2004, nitta:2014}.

When comparing complex-valued neural networks with real-valued neural networks, many publications ignore the number of parameters altogether \cite{aizenberg:2007}, compare only the number of parameters of the entire model \cite{trabelsi:2017}, or do not distinguish between complex- or real-valued parameters and units \cite{zhang:2017}. From the perspective of this paper such comparisons are equivalent to comparing models of different sizes. We systematically explore the performance of multi-layered perceptrons on simple classification tasks in consideration of the activation function, width and depth.

\section{Complex-Valued Neural Networks}
\label{complex_nn}
We define a complex-valued neuron analogous to its real-valued counter-part and consider its differences in structure and training. The complex neuron can be defined as:

\begin{equation}
\label{complex_neuron}
o = \phi (x \cdot w + b)
\end{equation}

with an activation function $\phi$ applied to the input $x \in \complexnumbers^{n}$, complex weight $w \in \complexnumbers^{n}$ and complex bias $ b \in \complexnumbers$. Arranging $m$ neurons into a layer:

\begin{equation}
\label{complex_dense}
o = \phi (x W + b)
\end{equation}

with an input $x \in \complexnumbers^{n}, W \in \complexnumbers^{n \times m}, b \in \complexnumbers^{m} $.

The activation function $\phi$ in the above definitions can be a function $\phi: \complexnumbers \to \realnumbers$ or $\phi: \complexnumbers \to \complexnumbers$. We will consider the choice of the non-linear activation function $\phi$ in more detail in Section \ref{activation}. In this work, we choose a simple real-valued loss function but complex-valued loss functions could be subject for future work. There is no total ordering on the field of complex numbers, since $i^2 = -1$. A complex-valued loss function would require defining a partial ordering on complex numbers (similar to a linear matrix inequality).

The training process in the complex domain differs, because activation functions are often not entirely complex-differentiable.

\begin{definition}
	\label{def:complex_diff}
	Analogous to a real function, a complex function $f: \complexnumbers \to \complexnumbers$ at a point $z_0$ of an open subset $\Omega  \subset \complexnumbers$ is \emph{complex-differentiable} if there exists a limit such that
	
	\begin{equation}
	\label{complex_diff}
	f'(z_0) = lim_{z \to z_0} \frac{f(z) -f(z_0)}{z -  z_0}
	\end{equation}
\end{definition}

If the function $f$ is complex-differentiable at all points of $\Omega$ it is called  \emph{holomorphic}. While in the real-valued case the existence of a limit is sufficient for a function to be differentiable, the complex definition in Equation \ref{complex_diff} implies a stronger property.

\begin{definition}
	\label{def:cauchy_riemann}
	A complex function $f(x + iy) = u(x, y) + i v(x,y)$ with real-differentiable functions $u(x,y)$ and $v(x,y)$ is complex-differentiable if they satisfy the Cauchy-Riemann Equations:
	
	\begin{equation}
	\begin{split}
	\frac{\partial u}{\partial x} = \frac{\partial v}{\partial y}, \quad
	- \frac{\partial u}{\partial y} = \frac{\partial v}{\partial x}
	\end{split}
	\label{cauchy_equation}
	\end{equation}
\end{definition}

We represent a complex number $z \in \complexnumbers$ with two real numbers $ z = x + i y$. For $f$ to be holomorphic, the limit not only needs to exist for the two functions $u(x,y)$ and $v(x,y)$, but the (partial) derivatives must also satisfy the Cauchy-Riemann Equations. That also means that a function can be non-holomorphic (i.e. not complex-differentiable) in $z$, but still be analytic in its parts $x, y$. Hence, to satisfy the Cauchy-Riemann Equations, real differentiability of functions $u(x,y)$ and $v(x,y)$ is not a sufficient condition to satisfy the Cauchy-Riemann Equations (Definition \ref{def:cauchy_riemann}).

In order to apply the \emph{chain rule} for non-holomorphic functions, the property of many non-holomorphic functions to be differentiable with respect to their real and imaginary parts can be utilised. We consider the complex function $f(z, \bar{z})$ to be a function of $z$ and its complex conjugate $\bar{z}$. Effectively, we choose a different basis for our partial derivatives.

\begin{equation}
\begin{split}
\frac{\partial}{\partial z} = \frac{1}{2} \big( \frac{\partial}{\partial x} - i \frac{\partial}{\partial y} \big),  \quad
\frac{\partial}{\partial \bar{z} } = \frac{1}{2} \big( \frac{\partial}{\partial x} + i \frac{\partial}{\partial y} \big)
\end{split}
\label{wirtinger_derivatives}
\end{equation}

These derivatives are a consequence of Wirtinger calculus (or $\complexnumbers \realnumbers$-calculus). They allow the application of the chain rule to many non-holomorphic functions for multiple complex variables $z_i$: 
\begin{equation}
\begin{split}
\label{complex_chainrule}
\frac{\partial}{\partial z_i} (f \circ g)& = \sum_{j=1}^{n} \big( \frac{\partial f}{\partial z_j} \circ g \big) \frac{\partial g_j}{\partial z_i} + \sum_{j=1}^{n} \big( \frac{\partial f}{\partial \bar{z}_j} \circ g \big) \frac{\partial \bar{g}_j}{\partial z_i}, \\
\frac{\partial}{\partial \bar{z}_i} (f \circ g)& = \sum_{j=1}^{n} \big( \frac{\partial f}{\partial z_j} \circ g \big) \frac{\partial g_j}{\partial \bar{z}_i} + \sum_{j=1}^{n} \big( \frac{\partial f}{\partial \bar{z}_j} \circ g \big) \frac{\partial \bar{g}_j}{\partial \bar{z}_i}
\end{split}
\end{equation}

Many non-holomorphic functions are also not entirely differentiable with respect to their real parts. The general practice of computing gradients only at specific points allows using a wide range of complex activation functions. The training process, however, can become numerically unstable. The unstable training process makes it necessary to devise special methods to avoid problematic regions of the function. The Wirtinger calculus, described above, provides an alternative method for computing the gradient that also improves the stability of the training process.

\section{Interaction of Parameters}
\label{interaction}

Any complex number $z = x + iy = r * e^{i \varphi}$ can be represented by two real numbers: the real part $Re(z) = x$ and the imaginary part $Im(z) = y$ or equivalently as magnitude $ |z| = \sqrt{x^2 + y^2} = r$ and phase (angle) $\varphi = arctan( \frac{x}{y})$. Consequently, any complex-valued function on one or more complex-variables can be expressed as a function on two real variables $f(z) = f(x, y) = f(r, \varphi)$.

Despite the straight forward use and representation in neural networks, complex numbers define an interaction between the two parts. Consider the operations necessary in the regression outlined in Equation \ref{complex_dense} to be composed of real and imaginary parts (or, equivalently, magnitude and phase). Each element $z_{1} \in \complexnumbers$ of the weight matrix $W \in \complexnumbers^{n \times m}$ interacts with an element $z_2 \in \complexnumbers$ of an input $x \in \complexnumbers^{n}$:

\begin{equation}
\begin{split}
	z_1 z_2& = (a+ib) (c+id) = (ac - bd) + i(ad + bc), \\
	z_1 + z_2& = (a+ib) + (c+id) = (a + c) + i(b + d)
\end{split}
\label{complex_op}
\end{equation}

In an equivalent representation with Euler's constant $e^{i\varphi} = cos(\varphi) + i sin(\varphi)$ as polar form.

\begin{equation}
\begin{split}
z_1 z_2& = (r_1  e^{i\varphi_1})(r_2  e^{i\varphi_2}) = (r_1 r_2  e^{i\varphi_1 + \varphi_2}),\\
z_1 + z_2& = (r_1  e^{i\varphi_1}) + (r_2  e^{i\varphi_2})\\& = r_1  cos(\varphi_1) + r_2  cos(\varphi_2) + i (r_2  sin(\varphi_2) + r_1  sin(\varphi_1))
\end{split}
\label{complex_op_euler}
\end{equation}

Complex parameters increase the computational complexity of a neural network as more operations are required. Instead of a single real-valued multiplication, up to four real multiplications and two real additions are required. As can be seen in Equations \ref{complex_op} and \ref{complex_op_euler} the computational complexity can be significantly reduced depending on the implementation and representation chosen.

Consequently, simply doubling the number of real-valued parameters per layer is not sufficient to achieve the same effect as in complex-valued neural networks. This is illustrated when a complex number $z = a + i b$ is expressed in a equivalent matrix representation. Specifically, as $2 \times 2$ matrix $M$ in the ring of $M_2(\realnumbers) $:

\begin{equation}
\begin{split}
	M_{(a+ib)}& = \begin{bmatrix}
	a & -b \\
	b & a  
	\end{bmatrix}
	\quad \text{such that} \quad M_{(0+i1)} = 
	\begin{bmatrix}
	0 & -1 \\
	1 & 0  
	\end{bmatrix}, \quad M_{(1+i0)} = 
	\begin{bmatrix}
	1 & 0 \\
	0 & 1  
	\end{bmatrix}
\end{split}
\end{equation}
\begin{equation}
\begin{split}
	M_{(a+ib)} M_{(c+id)}& = 
	\begin{bmatrix}
	a & -b \\
	b & a  
	\end{bmatrix}
	\begin{bmatrix}
	c & -d \\
	d & c  
	\end{bmatrix} = 
	\begin{bmatrix}
	ac - bd & bc + dc \\
	bc + dc & ac - bd
	\end{bmatrix}
\end{split}
\label{complex_mult_matrix}
\end{equation}

This \textit{augmented representation} facilitates computing the multiplication of an input $x$ with a complex-valued weight matrix $W$ as:
\begin{equation}
\begin{split}
	x W& = 
	\begin{bmatrix}
	Re(x)  &  -Im(x) \\
	Im(x)  &   Re(x)  
	\end{bmatrix}
	\begin{bmatrix}
	Re(W) \\
	Im(W)
	\end{bmatrix} = 
	\begin{bmatrix}
	Re(x)Re(W)-Im(x)Im(W) \\
	Im(x)Re(W)+Re(x)Im(W)
	\end{bmatrix}
\end{split}
\label{complex_matrix_mult}
\end{equation}

This interaction consequently means that architecture design needs to be reconsidered in order to facilitate this structure. A deep learning architecture that performs well with real-valued parameters may not work for complex-valued parameters and vice versa. Models that do not facilitate the structure or tasks that do not require complex-valued representations will not improve in performance.

Our experiments show that real-valued data does not require this structure. The imaginary part of the input $Im(x)$ is zero, so Equations \ref{complex_op} and \ref{complex_matrix_mult} simplify to:

\begin{equation}
\begin{split}
\label{complex_imaginary_zero}
	Re(x W) = Re(x)Re(W), \quad Im(x W) = Re(x)Im(W)
\end{split}
\end{equation}

For the training this means that the real parts $Re(x)$ and $Re(W)$ dominate the overall classification of a real-valued data point. In later sections we discuss our experiment results and illustrate the training with a synthetic classification task. 

\section{Capacity}
\label{capacity}
The number of (real-valued) parameters is a metric to quantify the capacity of a network in its ability to approximate structurally complex functions. With too many parameters the model tends to overfit the data while with too few parameters it tends to underfit.

A consequence of representing a complex number $a + ib$ using real numbers $(a,b)$ is that the number of real parameters of each layer is doubled: $p_{\complexnumbers} = 2 p_{\realnumbers}$. The number of real-valued parameters per layer should be equal (or at least as close as possible) between the real-valued and its complex-valued architecture. This ensures that models have the same capacity. Performance differences are caused by introducing complex numbers as parameters and not by a capacity difference.

Consider the number of parameters in a fully-connected layer in the real case and in the complex case. Let $n$ be the input dimension and $m$ the number of neurons, then the number of parameters of a real-valued layer $p_{\realnumbers}$ and of a complex layer  $p_{\complexnumbers}$ is given by

\begin{equation}
\begin{split}
	p_{\realnumbers} = (n \times m) + m, \quad
	p_{\complexnumbers} = 2(n \times m) + 2m
\end{split}
\label{params_dense}
\end{equation}

For a multi-layer perceptron with $k$ hidden layers, and output dimension $c$ the number of real-valued parameters without bias is given by:
\begin{equation}
\begin{split}
\label{params_mlp}
	p_{\realnumbers}& = n \times m + k(m \times m) + m \times c, \\
	p_{\complexnumbers}& = 2(n \times m) + 2k(m \times m) + 2(m  \times c)
\end{split}
\end{equation}

At first glance designing comparable multi-layer neural network architectures, i.e. with the same number of real-valued parameters in each layer, is trivial. However, halving the number of neurons in every layer will not achieve parameter comparability. The number of neurons define the output dimensions of a layer and the following layer's input dimension. We addressed this problem by choosing MLP architectures with an even number of hidden layers $k$ and the number of neurons per layer to be alternating between $m$ and $\frac{m}{2}$. We receive the same number of real parameters in each layer of a complex-valued MLP compared to a real-valued network. Let us consider the dimensions of outputs and weights with $k=4$ hidden layers. For the real-valued case:

\begin{equation}
\label{real_mlp4}
\begin{split}
	(1 \times n) \overbrace{(n \times m_1)}^{\text{Input layer}}
	&\to (1 \times m_1) \overbrace{(m_1 \times m_2)}^{\text{Hidden layer}} \\
	\to (1 \times m_2) \overbrace{(m_2 \times m_3)}^{\text{Hidden layer}}
	&\to (1 \times m_3) \overbrace{(m_3 \times m_4)}^{\text{Hidden layer}} \\
	\to (1 \times m_4) \overbrace{(m_4 \times m_5)}^{\text{Hidden layer}}
	&\to (1 \times m_5) \overbrace{(m_5 \times c)}^{\text{Output layer}} \\
	&\to \overbrace{(1 \times c)}^{\text{Model output}}
\end{split}
\end{equation}
where $m_i$ is the number of (complex or real) neurons of the $i$-th layer. The equivalent using $m_i$ complex-valued neurons would be:

\begin{equation}
\label{complex_mlp4}
\begin{split}
	(1 \times n) (n \times \frac{m_1}{2}) 
	&\to (1 \times \frac{m_1}{2})(\frac{m_1}{2} \times m_2) \\
	\to (1 \times m_2)(m_2 \times \frac{m_3}{2})
	&\to (1 \times \frac{m_3}{2})(\frac{m_3}{2} \times m_4) \\
	\to (1 \times m_4)(m_4 \times \frac{m_5}{2})
	&\to (1 \times \frac{m_5}{2})(\frac{m_5}{2} \times c) \\
	&\to (1 \times c)
\end{split}
\end{equation}

Another approach to the design of comparable architectures is to work with a parameter budget. Given a fixed budget of real parameters $p_{\realnumbers}$ we can define real or complex MLP with an even number $k \geq 0$ of hidden layers such that the network's parameters are within that budget. The $k$ hidden layers and the input layer have the same number of real or complex neurons $m_{\realnumbers} = m_{\complexnumbers}$. The number of neurons in the last layer is defined by the number of classes $c$.

\begin{equation}
\label{budget_mlp}
m_{\realnumbers} =
\begin{cases}
-\frac{n + c}{2k} + \sqrt{(\frac{n+c}{2k})^2  + \frac{p_{\realnumbers}}{k}}, & \text{if } k > 0  \\
\frac{p_{\realnumbers}}{n + c}, & \text{otherwise}
\end{cases}
\end{equation}

\begin{equation}
\label{budget_cmlp}
m_{\complexnumbers} =
\begin{cases}
-\frac{n + c}{2k} + \sqrt{(\frac{n+c}{2k})^2  + \frac{p_{\realnumbers}}{2k}}, & \text{if } k > 0  \\
\frac{p_{\realnumbers}}{2(n + c)}, & \text{otherwise}
\end{cases}
\end{equation}

\section{Activation Functions}
\label{activation}
In any neural network an important decision is the choice of non-linearity. With the same number of parameters in each layer, we are able to study the effects that activation functions have on the overall performance. An important theorem to be considered for the choice of activation function is the Liouville Theorem. The theorem states that any bounded holomorphic function $f: \complexnumbers \to \complexnumbers$ (that is differentiable on the entire complex plane) must be constant. Hence, we need to choose unbounded and/or non-holomorphic activation functions.

To investigate the performance of complex models assuming a function which is linearly separable in the complex parameters we chose the identity function. This allows us to identify tasks that may not be linearly separable in $\realnumbers$ using $m_{\realnumbers}$ neurons, but are linearly separable in $\complexnumbers$ using $m_{\complexnumbers}$ neurons. An example would be the approximation of the XOR function \cite{aizenberg:2016}. The hyperbolic tangent is a well-studied function and defined for both complex and real numbers. The rectifier linear unit is also well understood and frequently used in a real-valued setting, but has not been considered in a complex-valued setting. It illustrates separate application on the two parts of a complex number. The magnitude and squared magnitude functions are chosen to map complex numbers to real numbers.

\begin{itemize}
	\item Identity (or no activation function):
	\begin{equation}
	\phi(z) = z 
	\end{equation}
	
	\item Hyperbolic tangent:
	\begin{equation}
	\phi(z) = tanh(z) =  \frac{sinh(z)}{cosh(z)} = \frac{e^z - e^{-z}}{e^z + e^{-z}} = \frac{e^{2z} - 1}{e^{2z} + 1}
	\end{equation}
	
	\item Rectifier linear unit (ReLU):
	\begin{equation}
	\begin{split}
	\phi(z) = ReLU(z)& = ReLU\Big(Re(z)\Big) + ReLU\Big(Im(z)\Big)j \\&
	= max\Big(0, Re(z)\Big) + max\Big(0, Im(z)\Big) j
	\end{split}
	\end{equation}
	
	\item Intensity (or magnitude squared):
	\begin{equation}
	\phi(z) = |z|^2 = x^2 + y^2
	\end{equation}
	
	\item Magnitude (or complex absolute):
	\begin{equation}
	\phi(z) = |z| = \sqrt{x^2 + y^2}
	\end{equation}
\end{itemize}

Before applying the logistic function in the last layer we use another function $\phi: \complexnumbers \to \realnumbers$ to receive a real-valued loss. We chose the squared magnitude $\phi(z) = |z|^2 $. The intensity or probability amplitude of two interfering waves gives us a geometrically and probabilistically interpretable output.

\begin{equation}
sigmoid(|z|^2) = \frac{1}{1+e^{-x^2 - y^2}}
\end{equation}

For an output vector $z = [z_0, z_1, \dots, z_c]$
\begin{equation}
softmax(|z_j|^2) = \frac{e^{x^2 + y^2}}{\sum_{i=1}^{c} e^{x^2 + y^2}} 
\end{equation}

\section{Experiments}
\label{experiments}
\begin{figure}
	\centering
	\includegraphics[width=12.5cm]{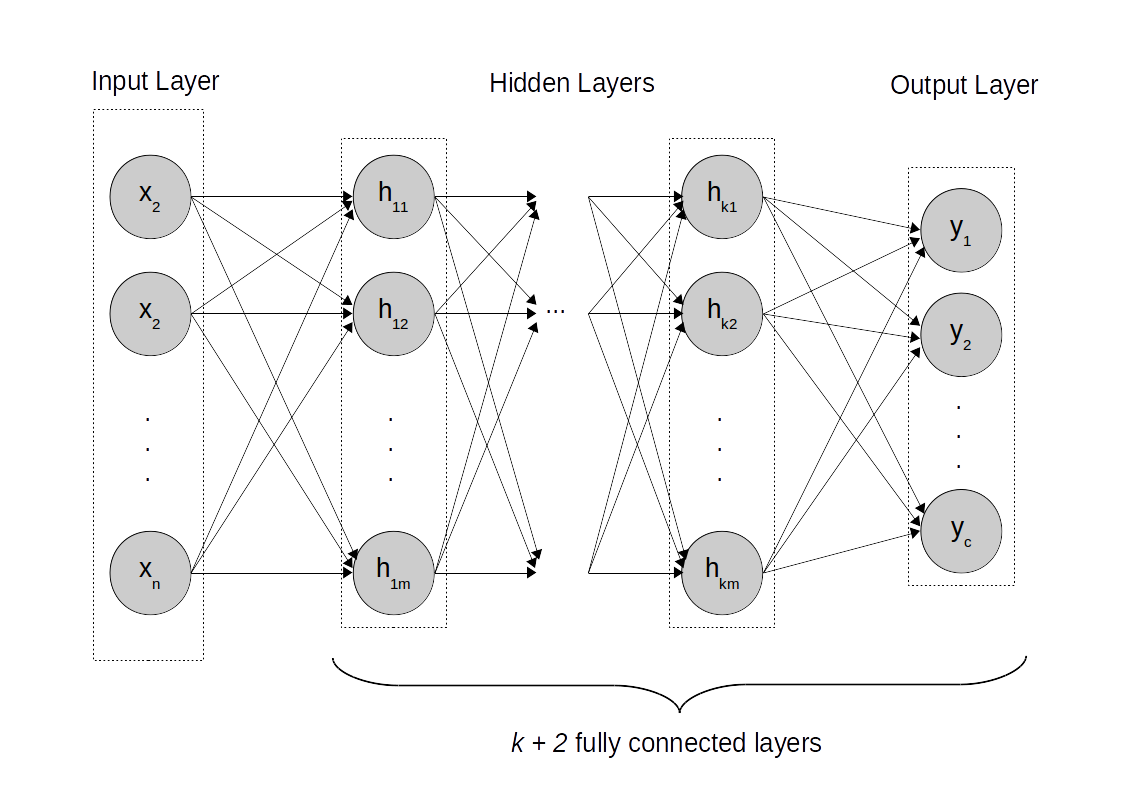}
	\caption{Multi-layer perceptron with $k$ hidden layers. The number of hidden units per layer $m$ computed with equation \ref{params_mlp} for Experiment 1 and Equations \ref{budget_mlp} resp. \ref{budget_cmlp} for Experiment 2.}
	\label{fig:mlp}
\end{figure}

To compare real and complex-valued multi-layer perceptrons (Figure \ref{fig:mlp}) we investigated them in various classification tasks. In all of the following experiments the task was to assign a single class to each real-valued data point using complex-valued multi-layer perceptrons:
\begin{itemize}
	
	\item Experiment 1: We tested MLPs with $k = 0, 2, 4, 8$ hidden layers, fixed width of units in each layer in real-valued architectures and alternating 64 and 32 units in complex-valued architectures (see section \ref{capacity}). We applied no fixed parameter budget. We tested the models on MNIST digit classification, CIFAR-10 Image classification, CIFAR-100 image classification and Reuters topic classification. Reuters topic classification and MNIST digit classification use $64$ units per layer, CIFAR-10 and CIFAR-100 use $128$ units per layer.
	
	\item Experiment 2: We tested MLPs with fixed budget of 500,000 real-valued parameters. The MLPs have variable width according to the depth and the parameters and are tested on MNIST digit classification, CIFAR-10 Image classification, CIFAR-100 image classification and Reuters topic classification. All tested activation functions are introduced in Section \ref{activation}. We rounded the units in Equations \ref{budget_mlp} and \ref{budget_cmlp} to the next integer.
\end{itemize}

We used the weight initialisation discussed by \citet{trabelsi:2017} for all our experiments. To reduce the impact of the initialisation we trained each model 10 times. Each run trained the model over 100 epochs with an Adam optimisation. We used categorical or binary cross entropy as a loss function depending on the task. We used $sigmoid(|z|^2)$ or $softmax(|z|^2)$ as the activation function for the last fully-connected layer.

\section{Results}
\label{results}
Tables \ref{table_mlp_mnist}, \ref{table_mlp_reuters}, \ref{table_mlp_cifar10},  \ref{table_mlp_cifar100} show the results for MLPs with variable depth and fixed width and no parameter budget (Experiment 1). Tables \ref{table_budget_mnist}, \ref{table_budget_reuters}, \ref{table_budget_cifar10}, \ref{table_budget_cifar100} show the results for MLPs with variable width according to depth and a fixed parameter budget of 500,000 real-valued parameters (experiment 2). In our experiments the achieved accuracy of complex- and real-valued multi-layer perceptrons are close to each other. Nevertheless, the real networks consistently outperform complex-valued networks and the complex-valued neural networks often fail to learn any structure from the data.

\begin{table}
	\caption{Test accuracy of a multi-layer perceptron consisting of $k + 2$ layers each with $64$ neurons (alternating $64$ and $32$ neurons in the complex MLP) and an output layer with $c=10$ neurons on MNIST digit classification task (Experiment 1). Selected best of 10 runs. Each run was trained for 100 epochs.}
	\begin{center}
		\label{table_mlp_mnist}
		\begin{tabular}{| c | c | c | c | c | }
			\hline
			\thead{Hidden\\layers\\k} & \thead{Real\\parameters\\$p_{\realnumbers}$} & \thead{Activation\\function\\$\varphi$} & \multicolumn{2}{c|}{MNIST} \\ \cline{4-5}
			&  &  & $\realnumbers$ & $\complexnumbers$ \\ \hline
			\multirow{5}{*}{k = 0} &  \multirow{5}{*}{50,816} & $identity$ & 0.9282 & 0.9509 \\ \cline{3-5}
			& & $tanh$ & 0.9761 & 0.9551  \\ \cline{3-5}
			& & $relu$ & 0.9780 & 0.9710  \\ \cline{3-5}
			& & $|z|^2$ & 0.9789 & 0.9609 \\ \cline{3-5}
			& & $|z| $ & 0.9770 & 0.9746  \\ \hline
			\multirow{5}{*}{k = 2} &  \multirow{5}{*}{59,008} & $identity$ & 0.9274 & 0.9482 \\ \cline{3-5}
			& & $tanh$ & 0.9795 & 0.8923 \\ \cline{3-5}
			& & $relu$ & 0.9804 & 0.9742  \\ \cline{3-5}
			& & $|z|^2$ & 0.9713 & 0.6573 \\ \cline{3-5}
			& & $|z| $ & 0.9804 & 0.9755 \\ \hline
			\multirow{5}{*}{k = 4} &  \multirow{5}{*}{67,200} & $identity$ & 0.9509 & 0.9468 \\ \cline{3-5}
			& & $tanh$ &  0.9802 & 0.2112 \\ \cline{3-5}
			& & $relu$ & 0.9816 & 0.9768 \\ \cline{3-5}
			& & $|z|^2$ & 0.8600 & 0.2572 \\ \cline{3-5}
			& & $|z| $ & 0.9789 & 0.9738 \\ \hline
			\multirow{5}{*}{k = 8} &  \multirow{5}{*}{83,584} & $identity$ & 0.9242 & 0.1771 \\ \cline{3-5}
			& & $tanh$ & 0.9796 & 0.1596 \\ \cline{3-5}
			& & $relu$ & 0.9798 & 0.9760 \\ \cline{3-5}
			& & $|z|^2$ &  0.0980 & 0.0980 \\ \cline{3-5}
			& & $|z| $ & 0.9794 & 0.1032 \\ \hline
		\end{tabular}
	\end{center}
\end{table}

\begin{table}
	\caption{Test accuracy of a multi-layer perceptron consisting of $k + 2$ layers each with $64$ neurons (alternating $64$ and $32$ neurons in the complex MLP) and an output layer with $c=46$ neurons on Reuters topic classification (Experiment 1). Selected best of 10 runs. Each run was trained for 100 epochs.}
	\begin{center}
		\label{table_mlp_reuters}
		\begin{tabular}{| c | c | c | c | c | }
			\hline
			\thead{Hidden\\layers\\k} & \thead{Real\\parameters\\$p_{\realnumbers}$} & \thead{Activation\\function\\$\varphi$} & \multicolumn{2}{c|}{Reuters} \\ \cline{4-5}
			&  &  & $\realnumbers$ & $\complexnumbers$ \\ \hline
			\multirow{5}{*}{k = 0} &  \multirow{5}{*}{642,944} & $identity$ & 0.8116 & 0.7939 \\ \cline{3-5}
			& & $tanh$ & 0.8117 & 0.7912  \\ \cline{3-5}
			& & $relu$ & 0.8081 & 0.7934  \\ \cline{3-5}
			& & $|z|^2$ & 0.8050 & 0.7885 \\ \cline{3-5}
			& & $|z| $ & 0.8068 & 0.7992  \\ \hline
			\multirow{5}{*}{k = 2} &  \multirow{5}{*}{651,136} & $identity$ & 0.8005 & 0.7836 \\ \cline{3-5}
			& & $tanh$ & 0.7978 & 0.7320 \\ \cline{3-5}
			& & $relu$ & 0.7921 & 0.7854  \\ \cline{3-5}
			& & $|z|^2$ & 0.7725 & 0.6874 \\ \cline{3-5}
			& & $|z| $ & 0.7996 & 0.7823 \\ \hline
			\multirow{5}{*}{k = 4} &  \multirow{5}{*}{659,328} & $identity$ & 0.7925 & 0.7787 \\ \cline{3-5}
			& & $tanh$ & 0.7814 & 0.4199 \\ \cline{3-5}
			& & $relu$ & 0.7734 & 0.7671 \\ \cline{3-5}
			& & $|z|^2$ & 0.5895 & 0.0650 \\ \cline{3-5}
			& & $|z| $ & 0.7863 & 0.7694 \\ \hline
			\multirow{5}{*}{k = 8} &  \multirow{5}{*}{675,712} & $identity$ & 0.7929 & 0.7796 \\ \cline{3-5}
			& & $tanh$ & 0.7542 & 0.1861 \\ \cline{3-5}
			& & $relu$ & 0.7555 & 0.7676 \\ \cline{3-5}
			& & $|z|^2$ & 0.0053 & 0.0053 \\ \cline{3-5}
			& & $|z| $ & 0.7671 & 0.7524 \\ \hline
		\end{tabular}
	\end{center}
\end{table}

\begin{table}
	\caption{Test accuracy of a multi-layer perceptron consisting of $k + 2$ layers each with $128$ neurons (alternating $128$ and $64$ neurons in the complex MLP) and an output layer with $c=10$ neurons on CIFAR-10 image classification task (Experiment 1). Selected best of 10 runs. Each run was trained for 100 epochs.}
	\begin{center}
		\label{table_mlp_cifar10}
		\begin{tabular}{| c | c | c | c | c | }
			\hline
			\thead{Hidden\\layers\\k} & \thead{Real\\parameters\\$p_{\realnumbers}$} & \thead{Activation\\function\\$\varphi$} & \multicolumn{2}{c|}{CIFAR-10} \\ \cline{4-5}
			&  &  & $\realnumbers$ & $\complexnumbers$ \\ \hline
			\multirow{5}{*}{k = 0} &  \multirow{5}{*}{394,496} & $identity$ & 0.4044 & 0.1063 \\ \cline{3-5}
			& & $tanh$ & 0.4885 & 0.1431  \\ \cline{3-5}
			& & $relu$ & 0.4902 & 0.4408  \\ \cline{3-5}
			& & $|z|^2$ & 0.5206 & 0.1000 \\ \cline{3-5}
			& & $|z| $ & 0.5256 & 0.1720  \\ \hline
			\multirow{5}{*}{k = 2} &  \multirow{5}{*}{427,264} & $identity$ & 0.4039 & 0.1000 \\ \cline{3-5}
			& & $tanh$ & 0.5049 & 0.1672 \\ \cline{3-5}
			& & $relu$ & 0.5188 & 0.496  \\ \cline{3-5}
			& & $|z|^2$ & 0.1451 & 0.1361 \\ \cline{3-5}
			& & $|z| $ & 0.5294 & 0.1000 \\ \hline
			\multirow{5}{*}{k = 4} &  \multirow{5}{*}{460,032} & $identity$ & 0.4049 & 0.1000 \\ \cline{3-5}
			& & $tanh$ & 0.4983 & 0.1549 \\ \cline{3-5}
			& & $relu$ & 0.8445 & 0.6810 \\ \cline{3-5}
			& & $|z|^2$ & 0.1000 & 0.1000 \\ \cline{3-5}
			& & $|z| $ & 0.5273 & 0.1000 \\ \hline
			\multirow{5}{*}{k = 8} &  \multirow{5}{*}{525,568} & $identity$ & 0.4005 &  0.1027 \\ \cline{3-5}
			& & $tanh$ & 0.4943 & 0.1365 \\ \cline{3-5}
			& & $relu$ & 0.5072 & 0.4939 \\ \cline{3-5}
			& & $|z|^2$ & 0.1000 & 0.1000 \\ \cline{3-5}
			& & $|z| $ & 0.5276 & 0.1000 \\ \hline
		\end{tabular}
	\end{center}
\end{table}

\begin{table}
	\caption{Test accuracy of a multi-layer perceptron consisting of $k + 2$ layers each with $128$ neurons (alternating $128$ and $64$ neurons in the complex MLP) and an output layer with $c=100$ neurons on CIFAR-100 image classification task (Experiment 1). Selected best of 10 runs. Each run was trained for 100 epochs.}
	\begin{center}
		\label{table_mlp_cifar100}
		\begin{tabular}{| c | c | c | c | c | }
			\hline
			\thead{Hidden\\layers\\k} & \thead{Real\\parameters\\$p_{\realnumbers}$} & \thead{Activation\\function\\$\varphi$} & \multicolumn{2}{c|}{CIFAR-100} \\ \cline{4-5}
			&  &  & $\realnumbers$ & $\complexnumbers$ \\ \hline
			\multirow{5}{*}{k = 0} &  \multirow{5}{*}{406,016} & $identity$ & 0.1758 & 0.0182 \\ \cline{3-5}
			& & $tanh$ & 0.2174 & 0.0142  \\ \cline{3-5}
			& & $relu$ & 0.1973 & 0.1793  \\ \cline{3-5}
			& & $|z|^2$ & 0.2314 & 0.0158 \\ \cline{3-5}
			& & $|z| $ & 0.2423 & 0.0235  \\ \hline
			\multirow{5}{*}{k = 2} &  \multirow{5}{*}{438,784} & $identity$ & 0.1720 & 0.0100 \\ \cline{3-5}
			& & $tanh$ & 0.2314 & 0.0146 \\ \cline{3-5}
			& & $relu$ & 0.2400 & 0.2123  \\ \cline{3-5}
			& & $|z|^2$ & 0.0143 & 0.0123 \\ \cline{3-5}
			& & $|z| $ & 0.2411 & 0.0100 \\ \hline
			\multirow{5}{*}{k = 4} &  \multirow{5}{*}{471,552} & $identity$ & 0.1685 & 0.0100 \\ \cline{3-5}
			& & $tanh$ & 0.2178 & 0.0157 \\ \cline{3-5}
			& & $relu$ & 0.2283 & 0.2059 \\ \cline{3-5}
			& & $|z|^2$ & 0.0109 & 0.0100 \\ \cline{3-5}
			& & $|z| $ & 0.2313 & 0.0100 \\ \hline
			\multirow{5}{*}{k = 8} &  \multirow{5}{*}{537,088} & $identity$ & 0.1677 & 0.0100 \\ \cline{3-5}
			& & $tanh$ &  0.2000 & 0.0130 \\ \cline{3-5}
			& & $relu$ & 0.2111 & 0.1956 \\ \cline{3-5}
			& & $|z|^2$ & 0.0100 & 0.0100 \\ \cline{3-5}
			& & $|z| $ & 0.2223 & 0.0100 \\ \hline
		\end{tabular} 
	\end{center}
\end{table}

\begin{table}
	\caption{Test accuracy of a multi-layer perceptron consisting of $k + 2$ dense layers with an overall budget of $500,000$ real-valued parameters on MNIST digit classification (experiment 2). The last layer consists of $c = 10$ neurons. Selected best of 10 runs. Each run was trained for 100 epochs.}
	\begin{center}
		\label{table_budget_mnist}
		\begin{tabular}{| c | c | c | c | c | c | }
			\hline
			\thead{Hidden\\layers\\k} &  \multicolumn{2}{c|}{Units} & \thead{Activation\\function\\$\varphi$} & \multicolumn{2}{c|}{CIFAR-10} \\ \cline{5-6}
			&  $m_{\realnumbers}$ & $m_{\complexnumbers}$ & & $\realnumbers$ & $\complexnumbers$ \\ \hline
			\multirow{5}{*}{k = 0} &  \multirow{5}{*}{630} & \multirow{5}{*}{315} & $identity$ & 0.9269 & 0.9464 \\ \cline{4-6}
			& & & $tanh$ & 0.9843 & 0.9467 \\ \cline{4-6}
			& & & $relu$ & 0.9846 & 0.9828 \\ \cline{4-6}
			& & & $|z|^2$ & 0.9843 & 0.9654 \\ \cline{4-6}
			& & & $|z| $ & 0.9857 & 0.9780 \\ \hline
			\multirow{5}{*}{k = 2} &  \multirow{5}{*}{339} & \multirow{5}{*}{207} & $identity$ & 0.9261 & 0.9427 \\ \cline{4-6}
			& & & $tanh$ & 0.9852 & 0.6608 \\ \cline{4-6}
			& & & $relu$ & 0.9878 & 0.9835 \\ \cline{4-6}
			& & & $|z|^2$ & 0.9738 & 0.8331  \\ \cline{4-6}
			& & & $|z| $ & 0.9852 & 0.9748  \\ \hline
			\multirow{5}{*}{k = 4} &  \multirow{5}{*}{268} & \multirow{5}{*}{170} & $identity$ & 0.9254 & 0.2943 \\ \cline{4-6}
			& & & $tanh$ & 0.9838 & 0.2002 \\ \cline{4-6}
			& & & $relu$ & 0.9862 &  0.9825 \\ \cline{4-6}
			& & & $|z|^2$ & 0.8895 & 0.2875 \\ \cline{4-6}
			& & & $|z| $ & 0.9846 & 0.9870 \\ \hline
			\multirow{5}{*}{k = 8} &  \multirow{5}{*}{205} & \multirow{5}{*}{134} & $identity$ & 0.9250 & 0.1136 \\ \cline{4-6}
			& & & $tanh$ & 0.9810 & 0.1682 \\ \cline{4-6}
			& & & $relu$ & 0.9851 & 0.9824 \\ \cline{4-6}
			& & & $|z|^2$ & 0.0980 & 0.0980 \\ \cline{4-6}
			& & & $|z| $ & 0.9803 & 0.1135 \\ \hline
		\end{tabular}
	\end{center}
\end{table}

\begin{table}
	\caption{Test accuracy of a multi-layer perceptron consisting of $k + 2$ dense layers with an overall budget of $500,000$ real-valued parameters on Reuters topic classification (experiment 2). The last layer consists of $c = 46$ neurons. Selected best of 10 runs. Each run was trained for 100 epochs.}
	\begin{center}
		\label{table_budget_reuters}
		\begin{tabular}{| c | c | c | c | c | c | }
			\hline
			\thead{Hidden\\layers\\k} &  \multicolumn{2}{c|}{Units} & \thead{Activation\\function\\$\varphi$} & \multicolumn{2}{c|}{Reuters} \\ \cline{5-6}
			&  $m_{\realnumbers}$ & $m_{\complexnumbers}$ & & $\realnumbers$ & $\complexnumbers$ \\ \hline
			\multirow{5}{*}{k = 0} &  \multirow{5}{*}{50} & \multirow{5}{*}{25} & $identity$ & 0.8072 & 0.7970 \\ \cline{4-6}
			& & & $tanh$ & 0.8112 & 0.7832 \\ \cline{4-6}
			& & & $relu$ & 0.8054 & 0.7925 \\ \cline{4-6}
			& & & $|z|^2$ & 0.8037 & 0.7929 \\ \cline{4-6}
			& & & $|z| $ & 0.8059 & 0.7912 \\ \hline
			\multirow{5}{*}{k = 2} &  \multirow{5}{*}{49} & \multirow{5}{*}{25} & $identity$ & 0.7992 & 0.7809 \\ \cline{4-6}
			& & & $tanh$ & 0.7952 & 0.7289 \\ \cline{4-6}
			& & & $relu$ & 0.7898 & 0.7751 \\ \cline{4-6}
			& & & $|z|^2$ & 0.7778 & 0.6887  \\ \cline{4-6}
			& & & $|z| $ & 0.7716 & 0.7911  \\ \hline
			\multirow{5}{*}{k = 4} &  \multirow{5}{*}{49} & \multirow{5}{*}{25} & $identity$ & 0.7636 & 0.7854 \\ \cline{4-6}
			& & & $tanh$ & 0.7796 & 0.4550 \\ \cline{4-6}
			& & & $relu$ & 0.7658 & 0.7676 \\ \cline{4-6}
			& & & $|z|^2$ & 0.5823 & 0.0289 \\ \cline{4-6}
			& & & $|z| $ & 0.7809 & 0.7573 \\ \hline
			\multirow{5}{*}{k = 8} &  \multirow{5}{*}{48} & \multirow{5}{*}{24} & $identity$ & 0.7760 & 0.7663 \\ \cline{4-6}
			& & & $tanh$ & 0.7449 & 0.1799 \\ \cline{4-6}
			& & & $relu$ & 0.7182 & 0.7484 \\ \cline{4-6}
			& & & $|z|^2$ & 0.0053 & 0.0053 \\ \cline{4-6}
			& & & $|z| $ & 0.7449 & 0.7302 \\ \hline
		\end{tabular}
	\end{center}
\end{table}

\begin{table}
	\caption{Test accuracy of a multi-layer perceptron consisting of $k + 2$ dense layers with an overall budget of $500,000$ real-valued parameters on CIFAR-10 image classification (experiment 2). The last layer consists of $c = 10$ neurons. Selected best of 10 runs. Each run was trained for 100 epochs.}
	\begin{center}
		\label{table_budget_cifar10}
		\begin{tabular}{| c | c | c | c | c | c | }
			\hline
			\thead{Hidden\\layers\\k} &  \multicolumn{2}{c|}{Units} & \thead{Activation\\function\\$\varphi$} & \multicolumn{2}{c|}{MNIST} \\ \cline{5-6}
			&  $m_{\realnumbers}$ & $m_{\complexnumbers}$ & & $\realnumbers$ & $\complexnumbers$ \\ \hline
			\multirow{5}{*}{k = 0} &  \multirow{5}{*}{162} & \multirow{5}{*}{81} & $identity$ & 0.4335 &  0.1006 \\ \cline{4-6}
			& & & $tanh$ & 0.5032 & 0.1676 \\ \cline{4-6}
			& & & $relu$ & 0.5007 & 0.4554 \\ \cline{4-6}
			& & & $|z|^2$ & 0.5179 & 0.1006 \\ \cline{4-6}
			& & & $|z| $ & 0.5263 & 0.2381 \\ \hline
			\multirow{5}{*}{k = 2} &  \multirow{5}{*}{148} & \multirow{5}{*}{77} & $identity$ & 0.4069 & 0.1000 \\ \cline{4-6}
			& & & $tanh$ & 0.5205 & 0.1673 \\ \cline{4-6}
			& & & $relu$ & 0.5269 & 0.4963 \\ \cline{4-6}
			& & & $|z|^2$ & 0.1395 & 0.1273  \\ \cline{4-6}
			& & & $|z| $ & 0.5315 & 0.1000  \\ \hline
			\multirow{5}{*}{k = 4} &  \multirow{5}{*}{138} & \multirow{5}{*}{74} & $identity$ & 0.4052 & 0.1000 \\ \cline{4-6}
			& & & $tanh$ & 0.5218 & 0.1475 \\ \cline{4-6}
			& & & $relu$ & 0.5203 & 0.4975 \\ \cline{4-6}
			& & & $|z|^2$ & 0.1065 & 0.1010 \\ \cline{4-6}
			& & & $|z| $ & 0.5234 & 0.1000 \\ \hline
			\multirow{5}{*}{k = 8} &  \multirow{5}{*}{123} & \multirow{5}{*}{69} & $identity$ & 0.4050 & 0.1003 \\ \cline{4-6}
			& & & $tanh$ & 0.5162 & 0.1396 \\ \cline{4-6}
			& & & $relu$ & 0.5088 & 0.4926 \\ \cline{4-6}
			& & & $|z|^2$ & 0.1000 & 0.1000 \\ \cline{4-6}
			& & & $|z| $ & 0.5194 & 0.1000 \\ \hline
		\end{tabular}
	\end{center}
\end{table}

\begin{table}
	\caption{Test accuracy of a multi-layer perceptron consisting of $k + 2$ dense layers with an overall budget of $500,000$ real-valued parameters on CIFAR-100 image classification (experiment 2). The last layer consists of $c = 100$ neurons. Selected best of 10 runs. Each run was trained for 100 epochs.}
	\begin{center}
		\label{table_budget_cifar100}
		\begin{tabular}{| c | c | c | c | c | c | }
			\hline
			\thead{Hidden\\layers\\k} &  \multicolumn{2}{c|}{Units} & \thead{Activation\\function\\$\varphi$} & \multicolumn{2}{c|}{CIFAR-100} \\ \cline{5-6}
			&  $m_{\realnumbers}$ & $m_{\complexnumbers}$ & & $\realnumbers$ & $\complexnumbers$ \\ \hline
			\multirow{5}{*}{k = 0} &  \multirow{5}{*}{158} & \multirow{5}{*}{79} & $identity$ & 0.2807 & 0.0314 \\ \cline{4-6}
			& & & $tanh$ & 0.2308 & 0.0193 \\ \cline{4-6}
			& & & $relu$ & 0.2153 & 0.1935 \\ \cline{4-6}
			& & & $|z|^2$ & 0.2364 & 0.0124 \\ \cline{4-6}
			& & & $|z| $ & 0.2439 & 0.0279 \\ \hline
			\multirow{5}{*}{k = 2} &  \multirow{5}{*}{144} & \multirow{5}{*}{75} & $identity$ & 0.1723 & 0.0100 \\ \cline{4-6}
			& & & $tanh$ & 0.2440 & 0.0203 \\ \cline{4-6}
			& & & $relu$ & 0.2481 & 0.2224 \\ \cline{4-6}
			& & & $|z|^2$ & 0.0155 & 0.0151  \\ \cline{4-6}
			& & & $|z| $ & 0.2453 & 0.0100  \\ \hline
			\multirow{5}{*}{k = 4} &  \multirow{5}{*}{135} & \multirow{5}{*}{72} & $identity$ & 0.1727 & 0.0100 \\ \cline{4-6}
			& & & $tanh$ & 0.2397 & 0.0150 \\ \cline{4-6}
			& & & $relu$ & 0.2381 & 0.2147 \\ \cline{4-6}
			& & & $|z|^2$ & 0.0122 & 0.0100 \\ \cline{4-6}
			& & & $|z| $ & 0.2390 & 0.0100 \\ \hline
			\multirow{5}{*}{k = 8} &  \multirow{5}{*}{121} & \multirow{5}{*}{67} & $identity$ & 0.1706 & 0.0100 \\ \cline{4-6}
			& & & $tanh$ & 0.2209 & 0.0164 \\ \cline{4-6}
			& & & $relu$ & 0.2167 & 0.2027 \\ \cline{4-6}
			& & & $|z|^2$ & 0.0100 & 0.0100 \\ \cline{4-6}
			& & & $|z| $ & 0.2191 & 0.0100 \\ \hline
		\end{tabular}
	\end{center}
\end{table}

Complex-valued MLPs can be used to classify short dependencies (e.g MNIST digit classification) or a short text as a bag-of-words (e.g. Reuters topic classification). For the two image classification tasks CIFAR-10 and CIFAR-100 the results indicate that a complex-valued MLP does not learn any structure in the data. These two tasks require larger weight matrices in the first layer and weight initialisation is still a significant problem.

The best non-linearity in complex neural network is the rectifier linear unit $relu$ applied to the imaginary and real parts, similarly to the real-valued models. $Identity$ and hyperbolic tangents $tanh$ outperform $relu$ - particularly in the real-valued case. However, the results using the rectifier linear unit $relu$ are much more stable. Despite the similarity of the activation functions $|z|^2$ and $|z|$, their performance in all tasks differ significantly. The magnitude $|z|$ consistently outperforms the squared magnitude $|z|^2$. In these classification benchmarks the activation function is the deciding factor for the overall performance of a given model. The activation may allow the network to recover from a bad initialisation and use the available parameters appropriately. An example would be the $relu$ activation in CIFAR task of Experiments 1 and 2 (Tables \ref{table_mlp_cifar10},  \ref{table_mlp_cifar100}, \ref{table_budget_cifar10}, \ref{table_budget_cifar100})

As expected, we observe that with a fixed number of neurons per layer (Experiment 1) and increasing depth, the complex- and real-valued accuracy increases. As we are increasing the total number of parameters, the model capacity increases. An exception here is Reuters topic classification where the performance decreases with increasing depth. When choosing the number of neurons per layer according to a given parameter budget (Experiment 2 using Equations \ref{budget_mlp}, \ref{budget_cmlp}), the performance decreases significantly as model depth increases. In consideration with the results from Experiment 1, the width of each layer is more important than the overall depth of the complete network.

We observed that the performance variance between the 10 initialisations is very high. We hypothesized that weight initialisation in complex MLPs becomes much more difficult with increasing depth. Hence, their performance is highly unstable. We confirmed this by training a complex MLP ($k=2$, $tanh$) with 100 runs (instead of 10 runs) on the Reuters classification task. The result shows a similar behaviour to the other results: The performance gap decreases if initialised more often. We found a test accuracy of 0.7748 in complex-valued case in comparison to 0.7978 in the real-valued case (Table \ref{table_mlp_reuters}). 

\section{Discussion}
\label{discussion}
For many applications that involve data that has an interpretation on the complex plane (e.g. signals) complex-valued neural networks have already shown that they are superior \cite{hirose:2009}. All selected tasks in our work use real-valued input data. We observe that complex-valued neural networks do not perform as well as expected for the selected tasks and real-valued architectures outperform their complex version. This finding seems counter-intuitive at first, since every real value is just a special case of a complex numbers with a zero imaginary part. Solving a real-valued problem with a complex-valued model allows the model greater degree of freedom to approximate the function. The question why complex-valued models are inferior to real models for the classification of real-valued data arises.

\begin{figure}
	\centering
	\includegraphics[width=14cm]{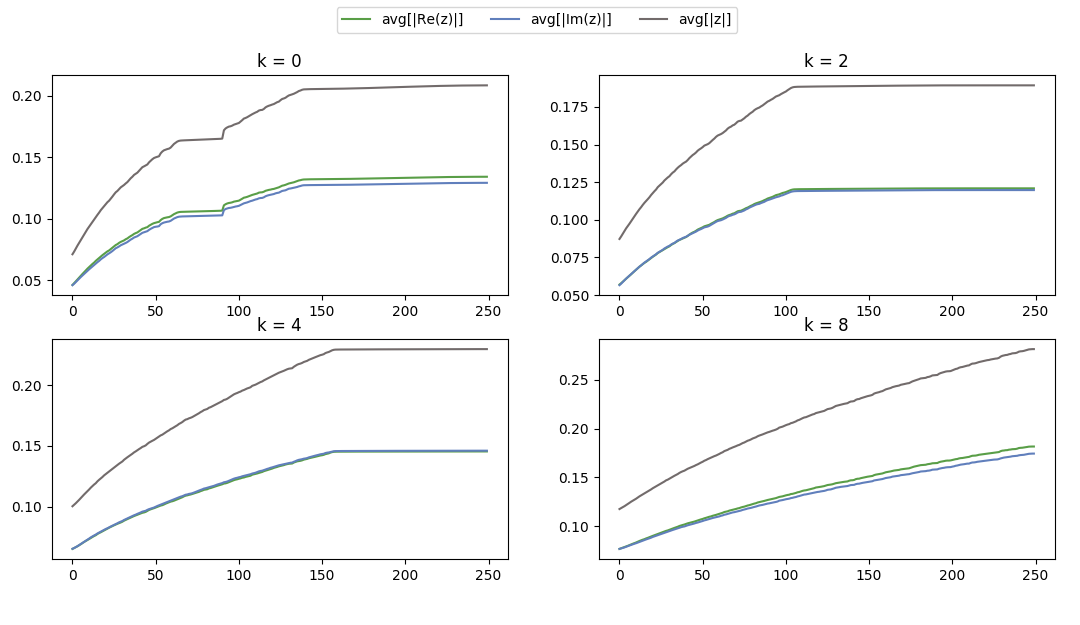}
	\caption{Average absolutes of real $|Re(z)|$, imaginary parts $|Im(z)|$ and average complex magnitude $|z|$ of all weights $W_i$ over training epochs of the MNIST classification task.}
	\label{fig:mnist_weight_behaviour}
\end{figure}

\begin{figure}
	\centering
	\includegraphics[width=14cm]{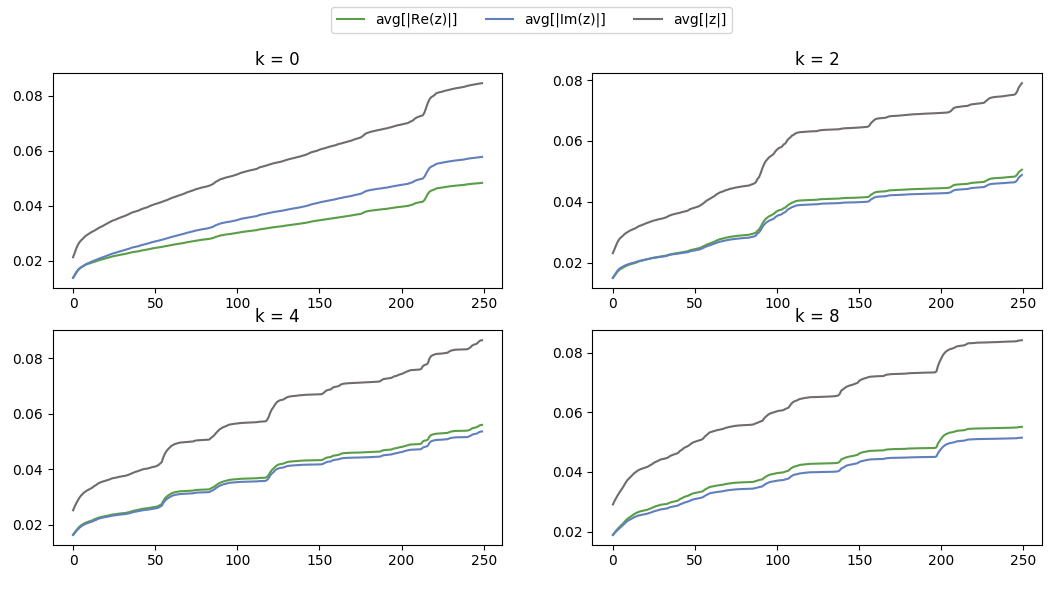}
	\caption{Average absolutes of real $|Re(z)|$, imaginary parts $|Im(z)|$ and average complex magnitude $|z|$ of all weights $W_i$ over training epochs of the Reuters classification task.}
	\label{fig:reuters_weight_behaviour}
\end{figure}

In further examination of the training process we observed that the imaginary parts of the complex weights always follow the real parts of the weights. We show this behaviour representatively with two tasks in Figure \ref{fig:mnist_weight_behaviour} and Figure \ref{fig:reuters_weight_behaviour}. 

\begin{figure}
	\centering
	\includegraphics[width=12.5cm]{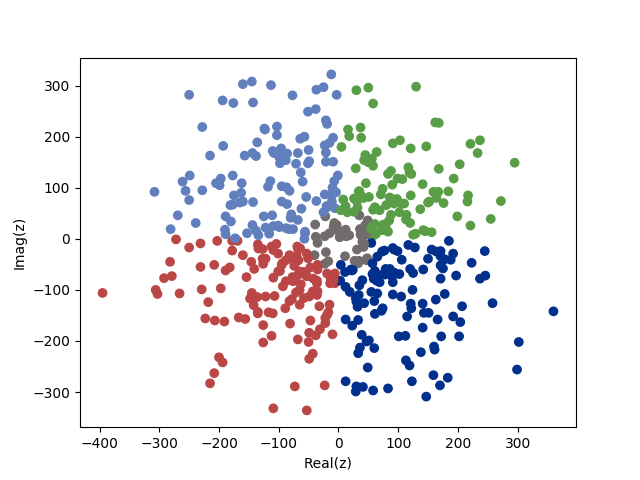}
	\caption{Example of synthetic data's class distribution. The class is determined by the location of the complex input vector's sum. Colours indicate different classes.}
	\label{fig:synthetic:complex_data}
\end{figure}

\begin{figure}
	\centering
	\includegraphics[width=12.5cm]{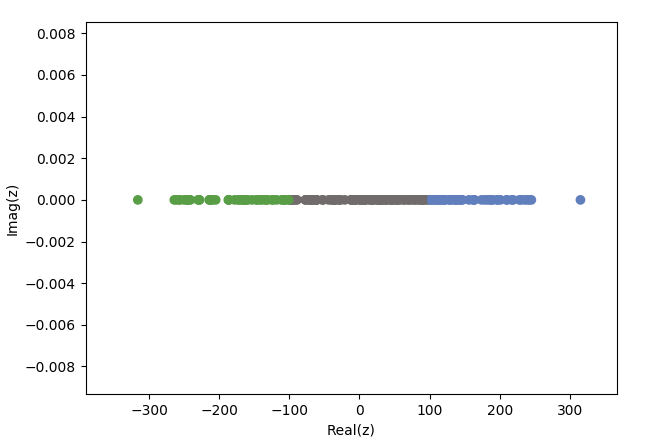}
	\caption{Example of synthetic data's class distribution. The class is determined by the location of the real input vector's sum. Colours indicate different classes.}
	\label{fig:synthetic:real_data}
\end{figure}

\begin{figure}
	\centering
	\includegraphics[width=12.5cm]{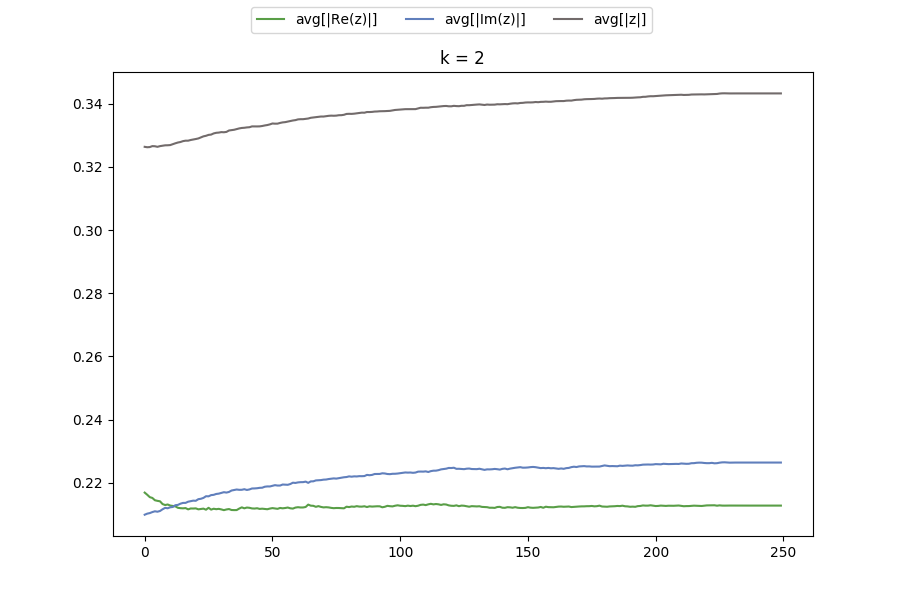}
	\caption{Example of independent learning behaviour of real and imaginary parts in the classification of synthetic complex-valued data. The real and imaginary parts change independently over the training. The exact trajectory of the graph depends on the weight initialisation.}
	\label{fig:synthetic:complex_behaviour}
\end{figure}

\begin{figure}
	\centering
	\includegraphics[width=12.5cm]{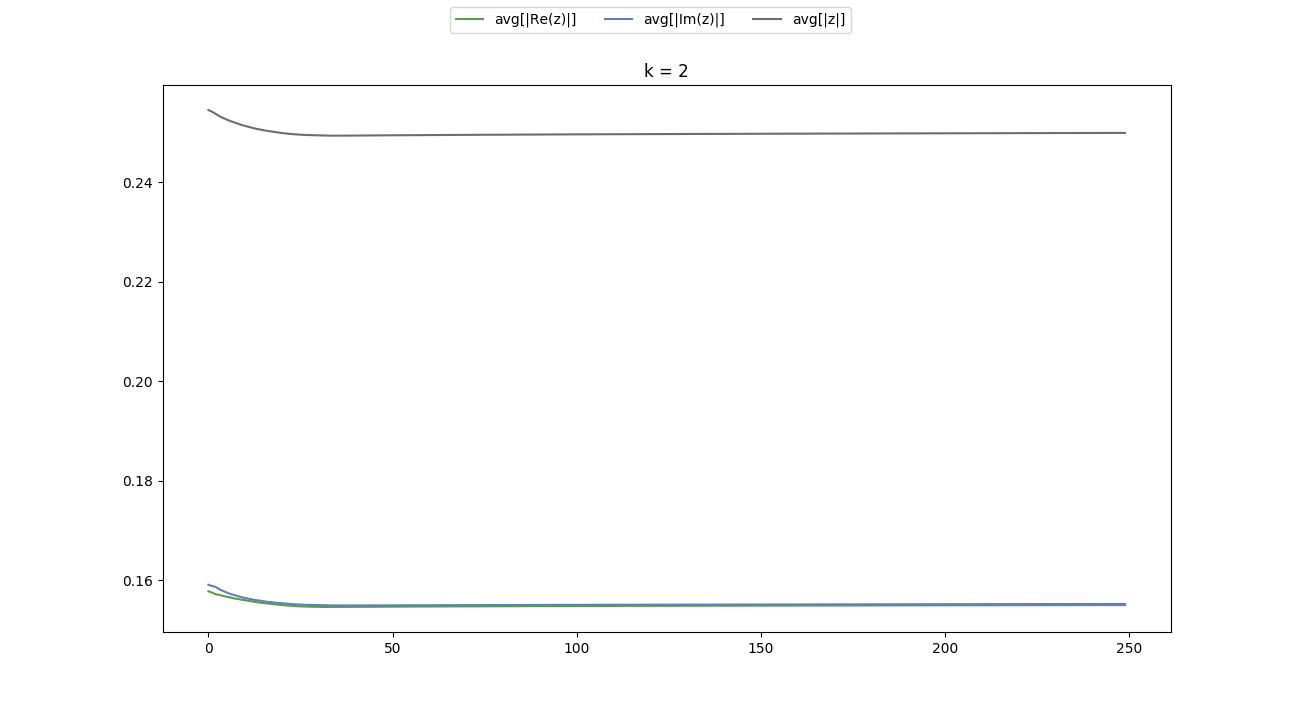}
	\caption{Example of dependent learning behaviour of the complex weights in the classification of synthetic real-valued data. The imaginary part follows the real part of the weight with every epoch. The exact trajectory of the graph depends on the weight initialisation.}
	\label{fig:synthetic:real_behaviour}
\end{figure}

We further investigate this behaviour with synthetic classification tasks in consideration of the information flow within a MLP. We create two synthetic classification task. Random complex data points $x \in \complexnumbers^{n}$ are to be classified using a complex-valued MLP according to the quadrant of its sum $\sum_{i=0}^{n} x_i$ or if it is close to the origin (Figure \ref{fig:synthetic:complex_data}). Real data points $x \in \realnumbers^{n}$ follow. This is equivalent to a projection of the complex data points to $\realnumbers$ (Figure \ref{fig:synthetic:real_data}). We classify $n = 10000$ complex resp. real input with Gaussian noise $\sigma = 0.2$ and $d = 25$ dimensions using a complex-valued MLP with $k=2$ hidden layers and with each $m=64$ units per layer.
Again, we observe that the weight initialisation is a significant problem. However, the complex model can reliably approximate the underlying complex or real functions achieving training accuracy of $0.981$, test accuracy of $0.908$. We observe that over the training process using complex input develops differently than for real input. Using complex input the real and imaginary parts develop independently and then reach convergence (Figure \ref{fig:synthetic:complex_behaviour}). Training with real-valued synthetic data the magnitudes of imaginary parts follow the real parts very closely (Figure \ref{fig:synthetic:real_behaviour}). Independently of the initialisation, the real parts converge a few epochs before the imaginary part (between 3 and 5 epochs).  We also tested non-linear and linear complex-valued regression approximating a real and a complex function. The imaginary part of the weights in a regression problem converges to zero or does not change if initialised with zero.

\begin{figure}
	\centering
	\includegraphics[width=14cm]{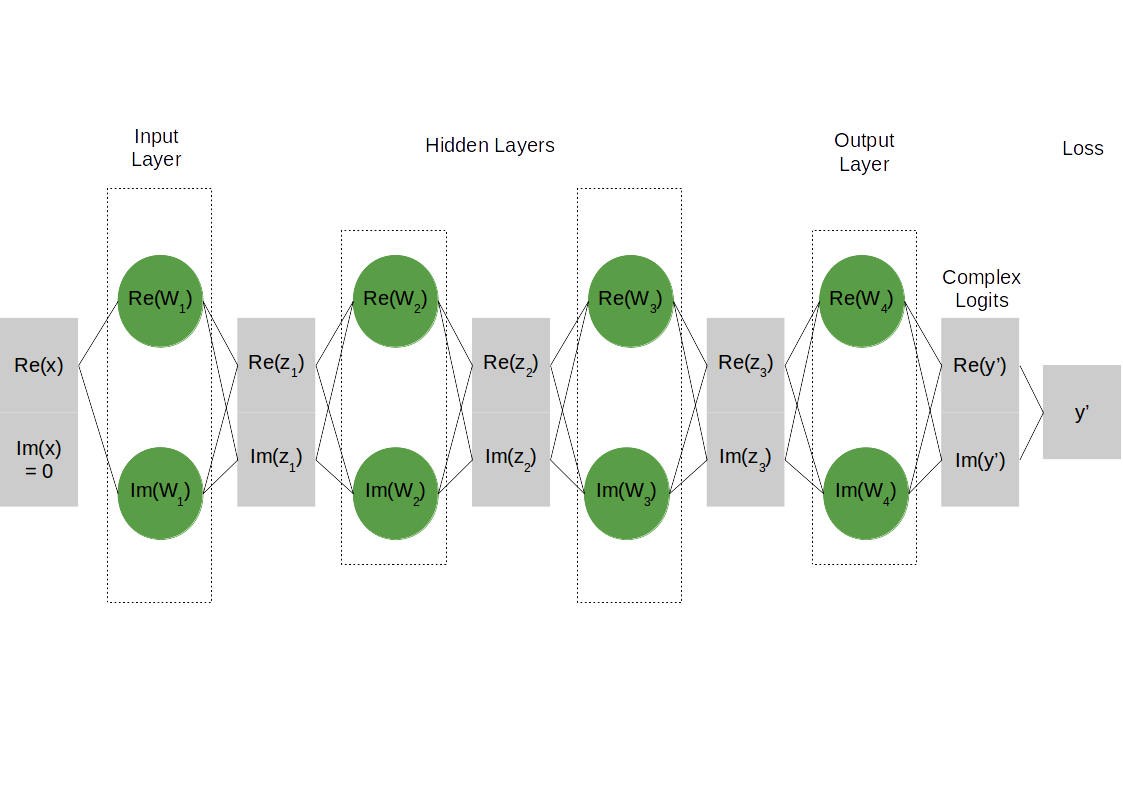}
	\caption{Interaction of real parts $Re(W_i)$, imaginary parts $Im(W_i)$ with the real-valued input $x$ and complex output $z_i$ of the $i$-th layer of an MLP with $k=2$ layers.}
	\label{fig:cmlp_information_flow}
\end{figure}

To explain the behaviour of the imaginary weights consider the computations of a complex-valued MLP combining two real weight matrices for a complex MLP in Figure \ref{fig:cmlp_information_flow}.

We see that the real and imaginary parts of the weights act identically on the input in order to reach a classification. The classification is thus the average of two identical classifications. If in the training phase the averaged absolute values of weight's imaginary part follow those of the real parts, the imaginary part of the input is either distributed exactly the same way as its real part, or the considered task simply does not benefit from using complex-valued hypothesis.

Moreover, we observed that complex-valued neural networks are much more sensitive towards their initialisation than real-valued neural networks. This sensitivity increases with the size of the network. The weight initialisation suggested by \citet{trabelsi:2017} can reduce this problem, but does not solve it. This initialisation method is a complex generalisation of the variance scaled initialisation by Glorot et. al. \cite{glorot:2010}. Other possible initialisations include the use of the random search algorithm (RSA) \cite{zimmermann:2011}. This requires significantly more computation. We eventually tried to mitigate the problem by running each experiment multiple times with different random initialisation. The initialisation of complex weights, however, is still a significant and unsolved problem and requires further investigation.

The unboundedness of the activation functions can cause numerical instability of the learning process. It can lead to a failed learning process (e.g. the gradient is practically infinite). If the learning process hits such a point in the function (e.g. a singularity), it is difficult to recover the training. It is avoidable by constraining a function, normalising weights or gradients. With increasing depth and structural complexity these options may be impractical due to their computational cost.  Alternatively, this can also be prevented in the design stage by choosing a bounded and entirely complex-differentiable activation function. Finding such a function is difficult. Another possibility is to avoid the problem in practice by applying separate bounded activation functions (the same or different real functions) help overcome this problem. The rectifier linear unit is one of these functions. While not entirely real-differentiable we found the training process is much more stable and the performance improved. Despite the mathematical difficulties due the differentiability, we can practically transfer a lot of insights from the real to the complex domain.

In summary real-valued models pose an upper performance limit for real-valued tasks when compared to complex-valued models of similar capacity, because the real and imaginary parts act identically on the input. An investigation of the information and gradient flow, can help to identify tasks that benefit from complex-valued neural networks.
In consideration of existing literature and our finding we recommend that complex neural networks should be used for classification tasks if the data is naturally in the complex domain, or can be meaningfully moved to the complex plane. The network should reflect the interaction of real and imaginary parts of weights with the input data. If the structure is ignored, the model may not be able to utilise the greater degree of freedom. It will most likely also require more initialisations and computational time due to the more complicated training process.

\section{Conclusion}
\label{conclusion}
This work considers a comparison between complex- and real-valued multi-layer perceptrons in benchmark classification tasks. We found that complex-valued MLPs perform similar or worse for classification of real-valued data, even if the complex-valued model allows larger degrees of freedom. We recommend the use of complex numbers in neural networks if a) the input data has a natural mapping to complex numbers, b) the noise in the input data is distributed on the complex plane or c) complex-valued embeddings can be learned from real-valued data. We can identify tasks that would benefit from it by comparing the training behaviour (e.g. by the average absolute values) of real and imaginary weights. If the imaginary part does not follow the real parts general behaviour across epochs, the task benefits from assuming a complex hypothesis.

Other aspects to consider for the model design are activation functions, weight initialisation strategies and the trade-off between performance, model size and computational costs. In our work, the best performing activation function is the component-wise application of the rectifier linear unit. We transfer many real and complex activation function by applying them separately on the two real parts, using Wirtinger Calculus or a gradient-based approach combined with strategies to avoid certain points. The initialisation described by \citet{trabelsi:2017} can help to reduce the initialisation problem, but further investigation is required. Similar to many other architectures, the introduction of complex numbers as parameters is also a decision to trade-off between the task-specific performance, the size of the model (i.e. the number of real-value parameters) and the computational cost.

\section*{Acknowledgements}

Nils M{\"o}nning was supported by the EPSRC via a Doctoral Training Grant (DTG) Studentship. Suresh Manandhar was supported by EPSRC grant EP/I037512/1, A Unified Model of Compositional \& Distributional Semantics:  Theory and Application.

\clearpage

\bibliographystyle{plainnat}
\bibliography{paper}

\end{document}